\documentclass[runningheads]{llncs}
\usepackage[T1]{fontenc}
\usepackage{multirow}

\usepackage{cite} % IEEE-compliant compressed numeric citations

\usepackage{textcomp}
\usepackage{bm}

\def\BibTeX{{\rm B\kern-.05em{\sc i\kern-.025em b}\kern-.08em
    T\kern-.1667em\lower.7ex\hbox{E}\kern-.125em X}}

\usepackage{mathtools}
\usepackage{amssymb,amsfonts}

\usepackage{graphicx}
\usepackage[caption=false,font=footnotesize]{subfig}

\usepackage{algorithm}
\usepackage{algorithmic}

\usepackage{booktabs}
\usepackage{array}
\usepackage{tabularx}
\usepackage{longtable}
\usepackage{ragged2e}

\usepackage{tikz}
\usetikzlibrary{arrows.meta,positioning,shapes.geometric}

\usepackage{enumerate}
\usepackage{pdflscape} % page rotation/landscape
\usepackage{url}       % simple URLs

\usepackage{hyperref}
\usepackage{cleveref} % keep after hyperref if hyperref is used

\usepackage{xcolor} % for \todo color

\newcommand{\orcid}[1]{\href{https://orcid.org/#1}{\textcolor[HTML]{A6CE39}{\aiOrcid}}}

\usepackage[acronym]{glossaries}     % Acronym handling (acronym list)
\makeglossaries
\glsdisablehyper

\usepackage{orcidlink}

\usepackage{booktabs}  % For professional-looking table lines (toprule, midrule, bottomrule)
\usepackage{tabularx}  % For tables that automatically fit to a specific width

\usepackage{lipsum}

\usepackage{longtable}
\usepackage{listings}
\definecolor{codegreen}{rgb}{0,0.6,0}
\definecolor{codegray}{rgb}{0.5,0.5,0.5}
\definecolor{codepurple}{rgb}{0.58,0,0.82}
\definecolor{backcolour}{rgb}{0.96,0.96,0.96}

\lstdefinestyle{scmstyle}{
    backgroundcolor=\color{backcolour},   
    commentstyle=\color{codegreen},
    keywordstyle=\color{magenta},
    numberstyle=\tiny\color{codegray},
    stringstyle=\color{codepurple},
    basicstyle=\ttfamily\footnotesize, % Use typewriter font
    breakatwhitespace=false,         
    breaklines=true,                 % Wraps long lines automatically
    captionpos=b,                    
    keepspaces=true,                 
    numbers=left,                    % Line numbers on the left
    numbersep=5pt,                  
    showspaces=false,                
    showstringspaces=false,
    showtabs=false,                  
    tabsize=2,
    frame=single,                    % Adds a frame around the code
    rulecolor=\color{gray!30}
}

\begin{document}
\title{From Gameplay Traces to Game Mechanics: Causal Induction with Large Language Models}
% \thanks{Supported by organization x.}}
%
% \titlerunning{Abbreviated paper title}
% If the paper title is too long for the running head, you can set
% an abbreviated paper title here
%
% \author{Mohit Jiwatode\inst{1,3}\orcidID{0009-0007-7980-6785} \and
% Alexander Dockhorn\inst{2}\orcidID{0000-0001-8711-7428} \and
% Bodo Rosenhahn\inst{1}\orcidID{0000-0003-3861-1424}}

\author{Mohit Jiwatode\inst{1,3}\and
Alexander Dockhorn\inst{2}\and
Bodo Rosenhahn\inst{1}}
\authorrunning{M. Jiwatode et al.}
\titlerunning{From Gameplay Traces to Game Mechanics}
% First names are abbreviated in the running head.
% If there are more than two authors, 'et al.' is used.
%
\institute{Institute for Information Processing, Leibniz University Hannover, Germany \and SDU Metaverse Lab, University of Southern Denmark, Denmark \and \email{jiwatode@tnt.uni-hannover.de} \\
}
\maketitle              % typeset the header of the contribution
\begin{abstract}
% \lipsum[1]

Deep learning agents can achieve high performance in complex game domains without often understanding the underlying causal game mechanics. To address this, we investigate Causal Induction: the ability to infer governing laws from observational data, by tasking Large Language Models (LLMs) with reverse-engineering Video Game Description Language (VGDL) rules from gameplay traces. To reduce redundancy, we select nine representative games from the General Video Game AI (GVGAI) framework using semantic embeddings and clustering. We compare two approaches to VGDL generation: direct code generation from observations, and a two-stage method that first infers a structural causal model (SCM) and then translates it into VGDL. Both approaches are evaluated across multiple prompting strategies and controlled context regimes, varying the amount and form of information provided to the model, from just raw gameplay observations to partial VGDL specifications. Results show that the SCM-based approach more often produces VGDL descriptions closer to the ground truth than direct generation, achieving preference win rates of up to 81\% in blind evaluations and yielding fewer logically inconsistent rules. These learned SCMs can be used for downstream use cases such as causal reinforcement learning, interpretable agents, and procedurally generating novel but logically consistent games.
\end{abstract}

\section{Introduction}

% Game AI and the needs of causal models
Game AI has achieved remarkable milestones in the last decade~\cite{mnih2015human, mathieu2023alphastar,silver2018general,risi2020chess}. Advances in Deep Reinforcement Learning (DRL) have enabled superhuman performance in combinatorially complex domains, from traditional board games like Go and Chess~\cite{silver2018general} to high-dimensional video games like Atari and StarCraft II~\cite{mnih2015human, mathieu2023alphastar}. However, these systems largely operate as ``black boxes'', relying only on statistical pattern recognition and learning correlations to master a policy that maximizes a reward. They rarely possess an explicit, transferable representation of the game's causal mechanics. The fundamental question---``\textit{Does the model understand \textit{why} an action leads to a specific outcome?}''---remains largely unanswered. This lack of interpretability creates significant hurdles: agents struggle to generalize when rules shift slightly, and their decision-making processes are opaque to human users~\cite{deng2023causal}.

To address this, we argue for moving beyond correlation-based learning towards causal reasoning. Rooted in the framework of Structural Causal Models (SCMs), causal AI offers the tools to reason about interventions (``\textit{What if I push this block?}'') and counterfactuals (``\textit{Would I have died if I hadn't jumped?}'')~\cite{pearl2009causality}. In this work, we propose the General Video Game AI (GVGAI) framework~\cite{gvgaibook2019} as a testbed for causal induction: the process of acquiring causal knowledge from experience~\cite{pearl2009causality}. GVGAI games are defined in the Video Game Description Language (VGDL)~\cite{schaul2014extensible}, a human-readable format that explicitly encodes game entities, interaction dynamics, initial state, and terminal conditions. We argue that VGDL is effectively a domain-specific Structural Causal Model; extracting it from raw observations is equivalent to discovering the causal laws of the game world.

While prior work has explored procedural content generation using Large Language Models (LLMs), we investigate a more fundamental capability: \textbf{Causal Induction}, i.e. ``\textit{Can an LLM observe a sequence of game states and reverse-engineer the underlying VGDL rule set?}'' We introduce a novel evaluation framework that tests this capability through two lenses: a \textbf{Multi-Class Game Identification} task to assess semantic understanding, and a \textbf{VGDL Synthesis} task comparing direct code generation against a neuro-symbolic, SCM-mediated approach.

\textbf{Our contributions are as follows:}
\begin{itemize}
    \item \textbf{Semantically Diverse Benchmark:} We construct a benchmark of 9 representative games from the GVGAI environment, identified via semantic clustering using S-BERT~\cite{SBERT} and K-Means~\cite{kmeans++} clustering to ensure rigorous testing across diverse game mechanics.
    \item \textbf{Self-Conditioned Evaluation:} We introduce a novel prompt-based evaluation for game identification that specifically tests an LLM's robustness against ``hallucinated'' or ``memorized'' mechanics versus actually understanding the game mechanics from observations.
    \item \textbf{SCM-Mediated Synthesis Framework:} We propose a dual-stream generation pipeline that forces LLMs to explicitly draft a causal graph (SCM) before generating code, evaluating whether intermediate causal reasoning improves the fidelity of synthesized game rules.
\end{itemize}

\noindent\textbf{Code repository:} \url{https://github.com/jiwatode-mohit/SCM_4_GVGAI}

% % -------------------------------------------------------------------
% %                        Background and related work
% % -------------------------------------------------------------------

\section{Background and Related Work}

\subsection{Causality and Structural Models}
Causality extends beyond statistical association, enabling systems to predict the effects of actions (interventions) and reason about alternative scenarios (counterfactuals). Judea Pearl~\cite{pearl2009causality} defines this hierarchy as the ``Ladder of Causation,'' distinguishing between mere association, intervention (altering a variable's value), and counterfactuals (reasoning about hypothetical alternatives). At the core of this framework are Structural Causal Models (SCMs), defined as a triple $(U, V, F)$, where endogenous variables $V$ are determined by exogenous variables and noise $U$ from structural functions $F$.

While Hammond et al.~\cite{hammond2023reasoning} have extended causal reasoning to game-theoretic models, empirical causal discovery---inferring $F$ from raw data---remains a challenge. In our work, we treat the game engine as the ground-truth SCM and the VGDL script as its symbolic representation. Unlike Baier et al.~\cite{baier2021causality}, who use causal subgoals for score optimization, our objective is the reconstruction of the causal laws themselves.

\subsection{VGDL as a Causal Proxy}
The GVGAI framework is widely utilized as a testbed for evaluating agent generalizability, challenging algorithms to adapt to diverse, real-time arcade environments~\cite{gvgaibook2019,balla2025multi}. The framework employs the Video Game Description Language (VGDL) to define these environments through four modular sections: a \texttt{SpriteSet} for instantiating distinct game actor types and static objects, a \texttt{LevelMapping} that translates ASCII symbols into spatial sprite instances, an \texttt{InteractionSet} for governing rule-based event handling, and a \texttt{TerminationSet} for establishing win/loss criteria. We propose that the VGDL schema provides a natural mapping to the Structural Causal Model (SCM) formalism. Specifically, we posit that the \texttt{SpriteSet} corresponds to the set of endogenous variables $V$, the \texttt{LevelMapping} defines the exogenous initialization context $U$, and the \texttt{InteractionSet} explicitly encodes the structural equations $F$ (e.g., \texttt{Avatar > Wall: stepBack}).

Prior attempts to generate VGDL have struggled with logical consistency. Nielsen et al.~\cite{nielsen2015towards} used evolutionary algorithms but found the resulting games often ``broken'' or unplayable due to incoherent mechanics. More recently, Hu et al.~\cite{hu2024game} demonstrated that while LLMs like GPT-4 can generate VGDL, they frequently ``hallucinate'' interactions that conflict with the game's physics. Our work addresses this by enforcing an intermediate SCM structure to ground the LLM's reasoning.

\subsection{Neurosymbolic Reasoning with LLMs}
While LLMs exhibit impressive fluency, they often lack robust logical reasoning and state tracking~\cite{hammond2023large, dave2024investigating}. To mitigate this, recent ``System 2'' approaches integrate LLMs with symbolic solvers~\cite{dutta2024frugal, pan2023logic} or use Chain-of-Thought (CoT) prompting~\cite{wei2022chain} to elicit intermediate reasoning steps.

In the domain of games, Todd et al. (GAVEL)~\cite{todd2024gavel} and Tanaka et al.~\cite{tanaka2024grammar} have explored using LLMs for Ludii game description language generation. However, Bateni et al.~\cite{bateni2025llm} highlight that models perform well on known games but fail on rule variations, reinforcing the need for a prompt setting to test whether a model is reasoning causally or simply retrieving memorized rules. Furthermore, Li et al.~\cite{li2024task} warn of ``task contamination,'' where zero-shot capabilities are often the result of memorization, a critical consideration when using public benchmarks like GVGAI.

% -------------------------------------------------------------------
%                        Methodology
% -------------------------------------------------------------------

\section{Methodology}

To systematically evaluate LLM capabilities within the GVGAI domain, we structured our methodology into four distinct phases: (1) the construction of a semantically diverse benchmark, (2) the definition of the experimental setup, including model and data selection, (3) a classification task to assess game identification, and (4) a synthesis task to evaluate rule induction and  code generation capabilities.

\subsection{Benchmark Construction}

The General Video Game AI (GVGAI) framework contains 116 distinct games, many of which are simple variants or share significant semantic overlap~\cite{gvgaibook2019}. To establish a robust evaluation baseline, we limited our pool to the 80 games that comprised the original training sets for the GVGAI competition\footnote{A snapshot of the training set is archived at: \url{https://web.archive.org/web/20190214020741/https://www.gvgai.net/training_set.php?rg=1}}.

Rather than selecting games randomly or subjectively, we sought to create a benchmark that maximizes semantic diversity. To achieve this, we developed a three-stage pipeline: translating game logic to natural language, embedding that language into a vector space, and clustering to identify a compact set of representative games.

% \subsubsection{Pipeline Component Selection}
% To select a model capable of high-quality summarization (for the pipeline), we first conducted a rapid preliminary classification test. For this initial assessment, we utilized a subset of 10 random games and their corresponding descriptions from a PhD thesis~\cite{dockhorn2020prediction}. The task was to predict the game from observations with a list of possible games and their descriptions. Qwen3-8B emerged as the best candidate LLM and was used for the VGDL translation task next.

% We evaluated varying models from cost-effective parameter categories (sub-4B and sub-8B) to ensure computational efficiency. As illustrated in Figure~\ref{fig:prelim_test}, the \textbf{Qwen3-8B} model emerged as a significant outlier, demonstrating superior performance compared to other tested architectures. Based on this result, we selected Qwen3-8B as the core engine for the VGDL translation task in our pipeline.

% \begin{figure}[ht]
%     \centering
%     \includegraphics[width=0.9\textwidth]{ICPR_2026_LaTeX_Templates/images/preliminary_test.png}
%     \caption{Preliminary accuracy assessment for a 10-way game classification task (using random games from~\cite{dockhorn2020prediction}). The Qwen3-8B model demonstrates significantly superior performance compared to other sub-4B and sub-8B models, justifying its selection for the VGDL translation task.}
%     \label{fig:prelim_test}
% \end{figure}

\subsubsection{Preliminary Model Selection}
To select a language model capable of high-quality summarization and further tasks, we first conducted a rapid preliminary classification test. For this initial assessment, we utilized a subset of 10 random games and their corresponding descriptions from a PhD thesis~\cite{dockhorn2020prediction}. LLMs were provided with gameplay traces along with a list of possible games and were tasked to identify which game generated these traces. Qwen3-8B emerged as the best candidate LLM, demonstrating superior performance amongst the tested cost-effective architectures (see Appendix), and was used for the VGDL translation task next.

\subsubsection{VGDL to Natural Language Translation}
The raw Video Game Description Language (VGDL) files consist of symbolic, high-level code. These files are not suitable for direct semantic embedding, as standard sentence embedding models are not trained on structured logical definitions~\cite{SBERT} but on natural language.

To bridge this modality gap between code and natural language, we ``translated'' the VGDL files for the 80 selected games into concise natural language descriptions ($<100$ words). Leveraging the results of our preliminary study, this task was performed using the \textbf{Qwen3-8B} model. The resulting descriptions provide a coherent natural language summary of the mechanics and goals of each game, suitable for embedding.

\subsubsection{Vector Embedding} The resulting natural language descriptions(for an example see~\Cref{app:vgdl_to_natural,app:vgdl} and  ) require an embedding model optimized for sentence-level semantics. We employed the Sentence-Transformer (S-BERT) framework~\cite{SBERT}, specifically the \texttt{all-MiniLM-L6-v2} model\cite{minilm}, to convert each game description into a 384-dimensional dense vector.

\subsubsection{Semantic Clustering}
To identify distinct semantic groups within this space, we apply the K-Means clustering algorithm~\cite{kmeans++} directly to the 384-dimensional embeddings. The optimal number of clusters was determined by performing a hyperparameter sweep for the cluster count $k$ ranging from 5 to 14. We conducted a comprehensive comparative analysis of various preprocessing strategies, including L2 normalization and Standard scaling~\cite{scikit-learn}, along with dimensionality reduction techniques like PCA~\cite{PCA} and UMAP~\cite{mcinnes2018umap}, which is detailed in Appendix~\ref{sec:appendix_clustering}. However, for our final benchmarks, we utilized the raw embeddings. This approach preserves the original semantic distances and aligns with the standard implementation demonstrated in the official Sentence-Transformers documentation~\cite{sbert_kmeans_example}. The optimal value of $k$ was found to be 9.

\begin{figure}[!htbp]
    \centering
    \includegraphics[width=0.8\textwidth]{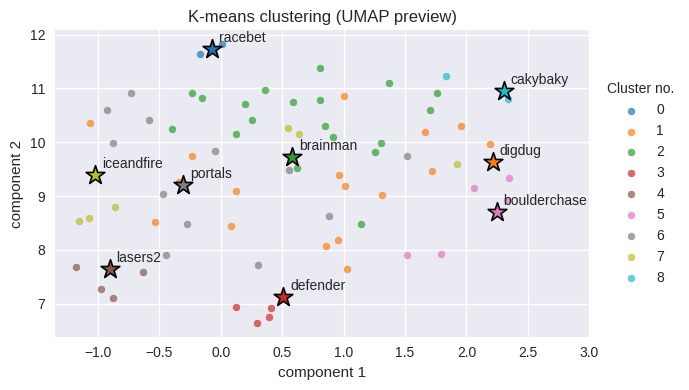}
    \caption{2D UMAP projection of the 80 games across 9 semantic clusters. The representative games are marked within their respective regions.}
    \label{fig:xyz}
\end{figure}

Consequently, we partitioned the 80 games into 9 distinct semantic clusters. The game closest to the centroid of each cluster was selected as the representative for that cluster, forming our 9-game benchmark. Figure~\ref{fig:xyz} provides a visual projection of these clusters.

% \todo{introduce the quantization of the models}

\subsection{Experimental Setup}

Following the identification of the 9 cluster representatives, we established a unified experimental environment for both classification and synthesis tasks.

\subsubsection{Model Selection}
While Qwen3-8B, a reasoning model, was selected for generation due to its preliminary success, we hypothesized that its performance might not be solely attributable to the architecture, but rather its underlying reasoning capabilities.

To investigate this, we expanded our model selection to include other models from the Qwen family to test architectural consistency, as well as other ``reasoning'' models along with their 4-bit quantized versions using Qlora~\cite{dettmers2023qlora}. This diverse selection allows us to decouple the effects of model architecture from general reasoning ability in the game identification task.

\subsubsection{Observation Generation}
% To generate the game observation sequences, we take random actions in the GVGAI framework. We collected data using 5 distinct random seeds to ensure robustness. For each seed, we extracted two types of sequences: the initial 10 frames of the game (\texttt{first\_10}) and the final 10 frames before the game concluded (\texttt{last\_10}), along with the corresponding random actions taken. This process yielded a total of 10 observation sets per game (5 seeds $\times$ 2 conditions). We limited the sequence length to 10 observations to constrain the context window, thereby preventing excessive GPU memory requirements and ensuring efficient processing speeds.

Ten sets of observation sequences per game were generated in the GVGAI framework, with each sequence limited to ten frames to constrain the context window and ensure computational efficiency.

\subsection{Task I: Multi-Class Game Identification}

%%%%%%%%%%%%%%%%%%%%%%%%%%%%%%%%%%%%%%%%%%%%%%%%%%%%%%%%%%%%%%%%%%%%%%%%%%%%%%%%%%%%%%%%%%%%%
%%%%%%%%%%%%%%%%%%%%%%%%%%%%%%%%%%%%%%%%%% OLD %%%%%%%%%%%%%%%%%%%%%%%%%%%%%%%%%%%%%%%%%%%%%%
%%%%%%%%%%%%%%%%%%%%%%%%%%%%%%%%%%%%%%%%%%%%%%%%%%%%%%%%%%%%%%%%%%%%%%%%%%%%%%%%%%%%%%%%%%%%%

% The first analytical task evaluates the capability of LLMs to correctly identify a game from the benchmark set based on ASCII-grid-based observations and a provided textual description. The LLM is provided with a prompt containing descriptions of the available games. It is then given a set of sequential ASCII-grid observations and must output a single class label.

% \subsubsection{Self-Conditioned Description Variants}
% Unlike standard classification benchmarks that use static prompts, we implemented a self-conditioned evaluation workflow. For each language model that is evaluated, we first task the model with generating descriptions for the games using three distinct strategies. These model-generated descriptions, along with expert descriptions ($P_{Standard}$), are then injected into the classification prompt. This measures the model's ability to classify observations based on its own internal understanding or summarization capabilities.

This task evaluates whether Large Language Models (LLMs) can identify a game from short sequences of ASCII-grid gameplay traces by reasoning about underlying mechanics, rather than relying on memorized names or prompt artifacts.

Given one set of candidate \textbf{game descriptions} (expert-authored or generated/refined by the model from the prompts below) and \textbf{gameplay traces} with actions, the model must output exactly one game label.

To isolate how different sources of semantic information influence classification, we evaluate models under four prompt variants. We refer to this setup as self-conditioned evaluation, as models are tested using descriptions they generate or refine themselves.

\begin{description}
    \item[$P_{Standard}$ (Human Expert):] Fixed expert-written descriptions focusing on mechanics and objectives. This serves as a high-quality semantic baseline. Unlike the other variants, this is a non-generative control condition. While it follows the same pipeline structure for simplicity, the LLM acts effectively as a pass-through, preserving the exact expert wording.

    \item[$P_{Cons}$ (Constructive Refinement):] The model refines the expert descriptions before classification. This tests whether models can improve causal clarity.

    \item[$P_{Dest}$ (Destructive/Parametric):] The model generates descriptions from the game name alone. This removes external context and controls for memorization~\cite{bateni2025llm} and task contamination~\cite{li2024task}.

    \item[$P_{VGDL}$ (VGDL-Summarization):] The model summarizes the raw ground-truth VGDLs into natural language descriptions. This tests the model's code-to-text reasoning.
\end{description}

\begin{figure}[htbp]
    \centering
    % Adjust width as needed, usually 0.8 to 1.0 of text width looks best
    \includegraphics[width=0.9\linewidth]{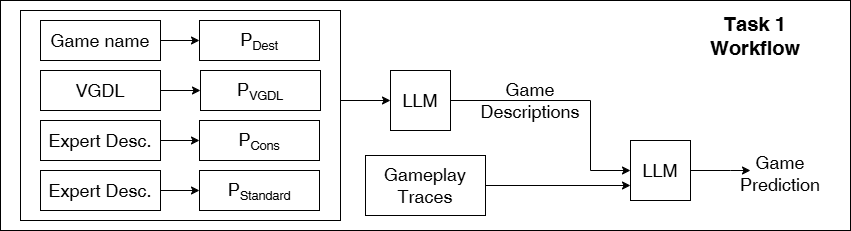} 
    \caption{\textbf{Task 1 Workflow:} The model predicts the game label based on a sequence of observations and a game description. The description is either fixed (Standard) or generated/refined by the LLM (Cons, Dest, VGDL) prior to classification.}
    \label{fig:task1_workflow}
\end{figure}

For each model, observation sequence, and prompt variant, we record the predicted label and classification accuracy. The complete workflow is illustrated in Figure~\ref{fig:task1_workflow}.

\subsection{Task II: VGDL Synthesis Framework}

The second analytical task utilizes a hierarchical context injection strategy to evaluate the capabilities of LLMs in synthesizing VGDL files directly from raw gameplay observations. The pipeline facilitates a comparative analysis between direct code synthesis and a causal-reasoning-mediated approach. 

\subsubsection{Efficiency Trade-off and Model Selection}

\begin{figure}[!htbp]
    \centering
    \includegraphics[width=0.9\textwidth]{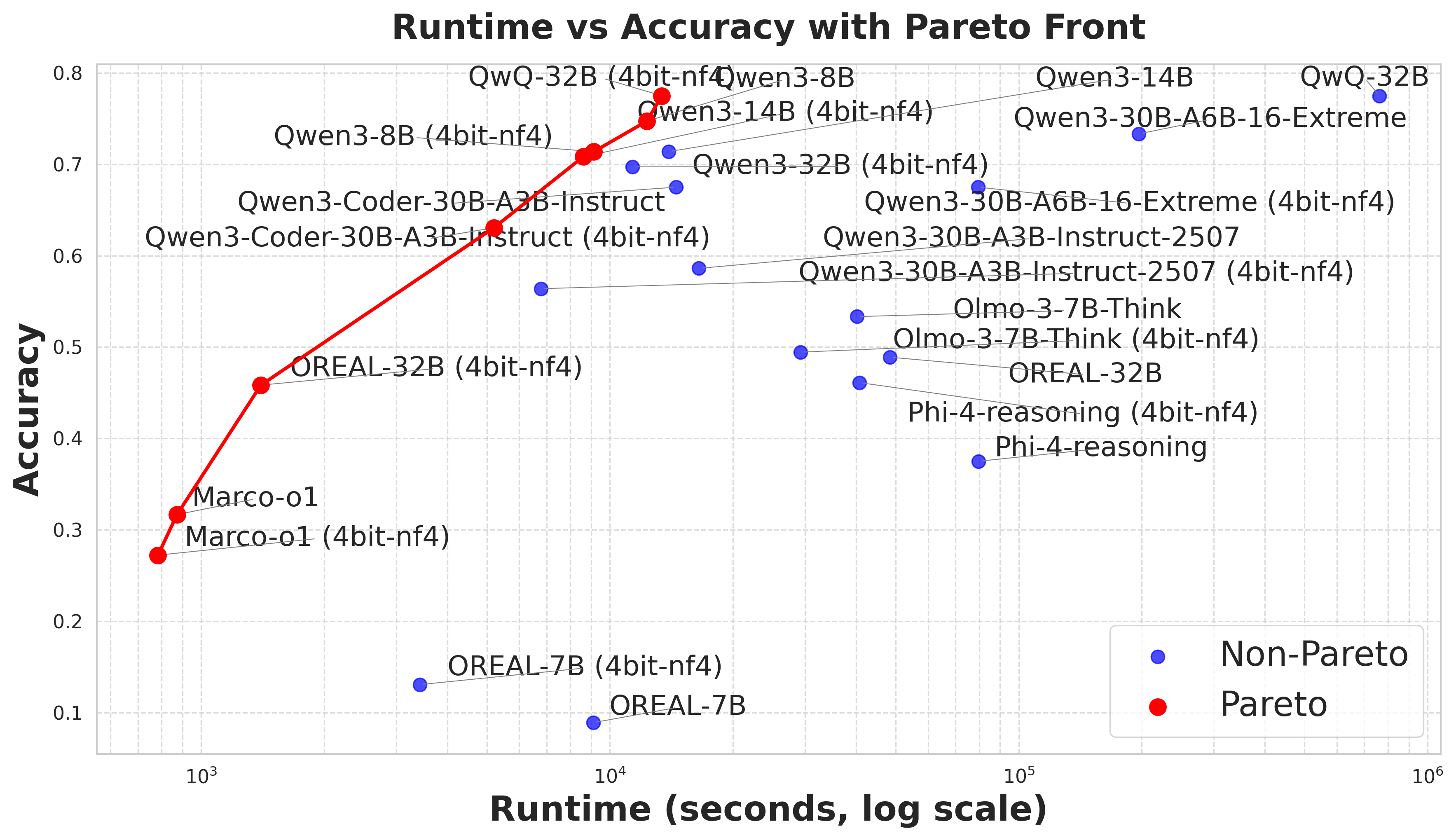}
    \caption{Runtime vs. Accuracy with Pareto Front. The red line indicates the Pareto frontier. Qwen3-8B and QwQ-32B lie on the frontier, justifying their selection for the resource-intensive synthesis task.}
    \label{fig:pareto}
\end{figure}

To identify models suitable for this computationally intensive VGDL synthesis, we analyzed the trade-off between accuracy and logarithmic runtime (Figure~\ref{fig:pareto}) for Task I. From the \textbf{Pareto Front} (red line), we selected two candidates that exceed 75\% accuracy, balancing performance with feasibility: \textbf{Qwen3-8B} and \textbf{QwQ-32B (quantized)}.

\subsubsection{Hierarchical Context Engineering}
To isolate the impact of information density on code generation, we define five cumulative levels of context ($C_L$). This extends the approach of Hu et al.~\cite{hu2024game} of using progressively informative prompt contexts to systematically isolate the effects of information. At each level, the model is tasked with generating a complete VGDL definition.

\begin{itemize}
    \item \textbf{Level 0 (The Baseline):} The model is provided strictly with the raw grid observation string (ASCII matrices of the game state). No semantic descriptions, grammar rules, or game names are included. This tests if the model can infer physics dynamics solely from visual state transitions.

    \item \textbf{Level 1 (The Syntax Check):} The prompt is augmented with the \textit{Common VGDL Grammar} specification and a single example of a valid VGDL file from a game not in the benchmark set of games we test on. This tests if providing valid vocabulary (e.g., \texttt{SpriteSet}, \texttt{InteractionSet}) allows the model to structure its physical inferences into valid code?

    \item \textbf{Level 2 (The Exact Game):} The context includes the syntax (Level 1) plus the target game's name and its specific high-level natural language description. This tests if the model can correlate natural language rules (e.g., ``The avatar must avoid the goblin'') with the visual grid entities.

    \item \textbf{Level 3 (The Distractor Test):} The input includes the observation and syntax, but instead of the single correct description, the model is provided with a dictionary of potential games (the target game plus distractors). This tests if the model can implicitly identify the correct game mechanics when presented with competing rule sets.

    \item \textbf{Level 4 (The Completionist):} The model is provided with a partial VGDL file where specific logic blocks have been removed. Specifically, the \texttt{SpriteSet} (entity definitions) and \texttt{LevelMapping} (layout) are preserved, while the \texttt{InteractionSet} (collision physics) and \texttt{TerminationSet} (win conditions) are completely removed. This tests if the model can reverse-engineer the missing interaction logic given the static entity definitions and visual evidence.
\end{itemize}

\subsubsection{Dual-Stream Generation Architecture}
For every context level $L$, the framework executes two parallel generation tasks to produce the final VGDL output. The inputs and constraints differ as follows:

\begin{itemize}
    \item \textbf{Stream A: Direct Synthesis (Non-SCM):}
    \begin{itemize}
        \item The input is the JSON-structured payload containing the accumulated context for Level $L$.
        \item The model acts as a ``VGDL Architect.'' It is instructed to perform an analysis of the objects in the grid before directly outputting the raw, executable VGDL code block derived from pattern matching and context inference.
    \end{itemize}
    
    \item \textbf{Stream B: Structural Causal Model (SCM) Guided:}
    \begin{itemize}
        \item This stream utilizes the same context Level $L$ but enforces a constrained Chain-of-Thought (CoT) approach defined by a strict \textbf{SCM Blueprint}. This approach is directly motivated by the findings of Hu et al.~\cite{hu2024game}, who observed that LLMs frequently ``hallucinate'' incorrect game logic due to conflicts with natural language word order. Furthermore, Tanaka et al.~\cite{tanaka2025grammar} demonstrated that grammar constraints alone are insufficient for capturing correct mechanics without gameplay alignment. By enforcing an intermediate SCM representation, we provide the structured reasoning aid that Bateni et al.~\cite{bateni2025rule} identified as necessary for accurate rule simulation.
        \item \textbf{Step 1 (Causal Graph Generation):} The model first generates a structural causal model in JSON format. The graph nodes are strictly typed into layers: \textit{Design} (e.g., ActionSpace), \textit{Dynamics} (e.g., InteractionMechanics), and \textit{Observation} (e.g., LevelEncoding).
        \item \textbf{Step 2 (Compiler Phase):} The generated SCM JSON is fed back into the model with instructions to ``compile'' the causal graph into executable VGDL code.
    \end{itemize}
\end{itemize}

\begin{figure}[htbp]
    \centering
    \includegraphics[width=0.8\linewidth]{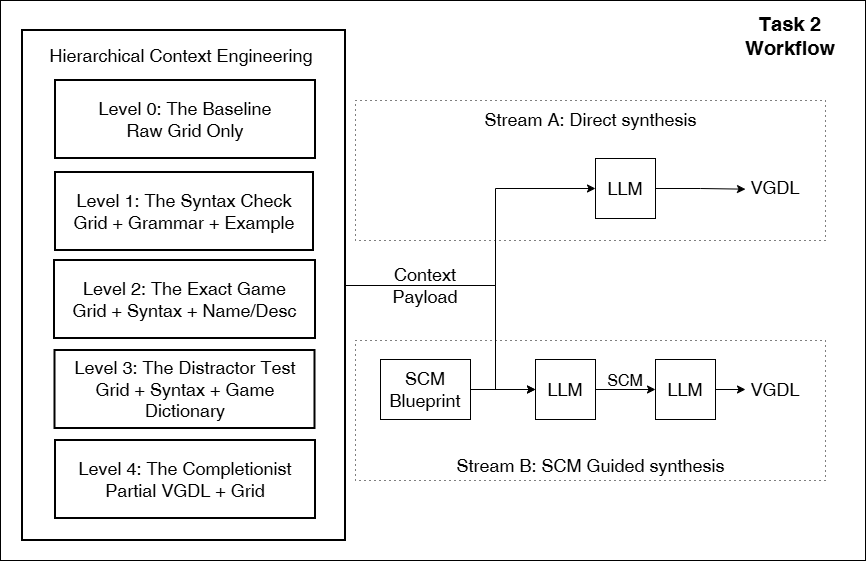}
    
    \caption{\textbf{Task II Workflow Diagram.} This overview illustrates the hierarchical context engineering levels (left) feeding into the dual-stream generation architecture (right). Stream A utilizes direct synthesis, while Stream B enforces a Structural Causal Model (SCM) blueprint.}
    
    \label{fig:task2_workflow}
\end{figure}

As illustrated in Figure~\ref{fig:task2_workflow}, the selected context payload is processed through two parallel streams: a direct synthesis approach (Stream A) and a causal-reasoning-mediated approach (Stream B).

\subsubsection{Evaluation Metrics}
The fidelity of the generated VGDL is assessed in a post-processing phase using three distinct metrics:

\begin{enumerate}
    \item \textbf{VGDL Cosine Similarity:} The cosine similarity between the vector embeddings of the generated VGDL code and the ground truth VGDL file.
    \item \textbf{Semantic Text Similarity:} The cosine similarity between the natural language descriptions derived by translating both the generated and ground truth VGDL files using Qwen3-8B, assessing conceptual alignment.
    \item \textbf{LLM-based Preference Evaluation:} To capture qualitative alignment, three state-of-the-art closed-source models (GPT-5.2, Gemini 2.5 Pro, Claude 4.5 Opus), and the Qwen3-8B model are presented with the ground truth VGDL and the outputs from Stream A and Stream B. They are queried to perform a blind ranking of which stream's output better preserves the semantic and logical integrity of the original game mechanics from the ground truth VGDL.
\end{enumerate}

% -------------------------------------------------------------------
%                        Results, discussion
% -------------------------------------------------------------------

\section{Results and Discussion}
\label{sec:results}

\subsection{Task I: Multi-Class Game Identification}

We evaluated a suite of Large Language Models (LLMs) on their ability to identify games from ASCII grid observations across four distinct description prompt types ($P_{VGDL}, P_{Standard}, P_{Cons}, P_{Dest}$).

\subsubsection{Overall Model Performance}
\begin{figure}[!htbp]
    \centering
    \includegraphics[width=\textwidth]{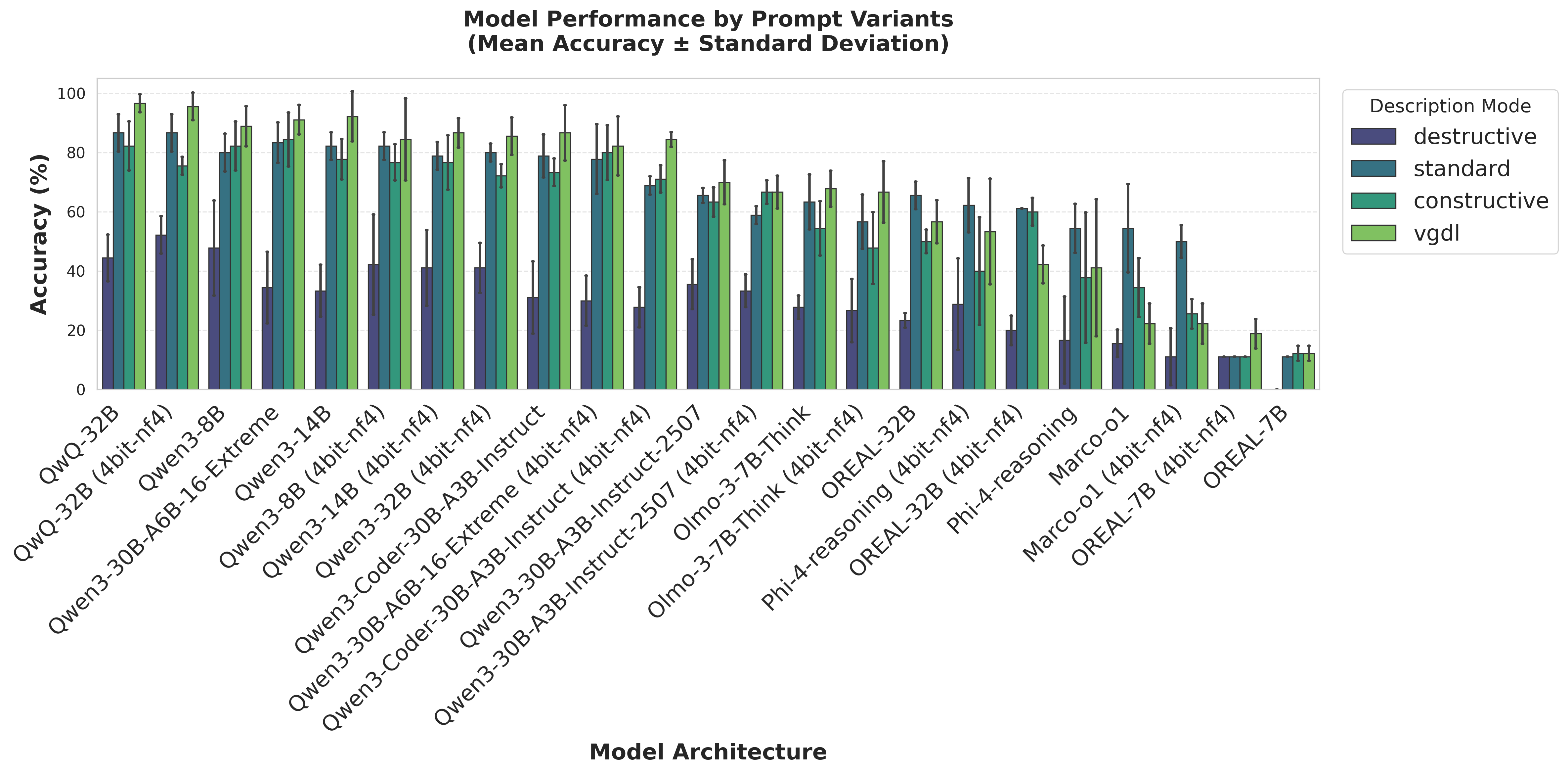}
    \caption{Model Performance by Description Mode. The figure illustrates the mean accuracy across description types. The stark drop in the $P_{Dest}$ trajectory compared to $P_{Standard}$ indicates that models lack complete internal knowledge of the games, relying heavily on provided context.}
    \label{fig:performance_mode}
\end{figure}

The overall classification accuracy varied significantly across model architectures. As shown in Figure~\ref{fig:performance_mode}, the \textbf{QwQ-32B (quantized)} model achieved the highest global average accuracy ($\approx 77.5\%$) with a runtime of around 13300 seconds, followed closely by \textbf{Qwen3-8B} ($\approx 74.7\%$) at a slightly lower runtime of around 12300 seconds. In contrast, other families of models, such as OREAL-7B and Marco-01, yielded near-random performance, highlighting the limitation of these models for understanding ASCII-based spatial reasoning.

Figure~\ref{fig:performance_mode} further illustrates a critical finding regarding the models' internal knowledge. We observe a \textbf{stark drop in performance} when moving from the context-rich settings ($P_{Standard}$ or $P_{VGDL}$) to the Destructive setting ($P_{Dest}$), where the model must rely solely on its training data to generate the game description.
\begin{itemize}
    \item \textbf{Context Dependency:} For top-performing models like Qwen3-8B, the accuracy drops from approximately 80\% in the Standard setting to roughly 45\% in the Destructive setting. This degradation supports the hypothesis presented in our Methodology: despite their size, these models do not possess a complete or robust understanding of the specific GVGAI game rules from their training data alone.
    \item \textbf{Validation of Methodology:} The inability of models to maintain performance without external descriptions validates our use of the Destructive prompt as a control variable. It confirms that the models do not have a good understanding of the game mechanics.
\end{itemize}

\subsubsection{Semantic Confusion Analysis}
\begin{figure}[!htbp]
    \centering
    \includegraphics[width=0.65\textwidth]{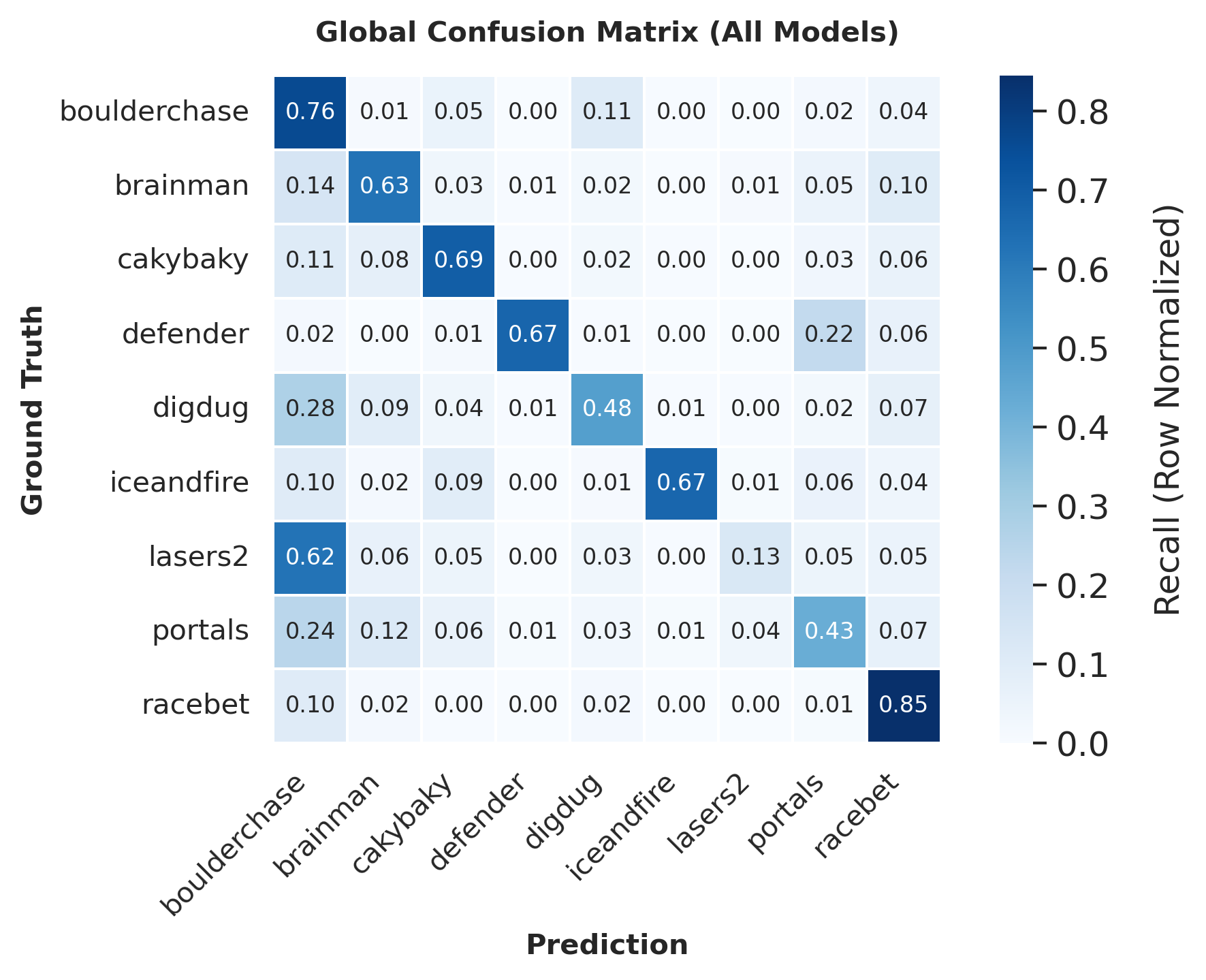}
    \caption{Global Confusion Matrix aggregated across all models. High diagonal values indicate effective identification of archetypal games, though semantic bleed remains between mechanically similar titles like \textit{digdug} and \textit{boulderchase}.}
    \label{fig:confusion}
\end{figure}

To validate the effectiveness of our semantic clustering, we analyzed the global confusion matrix (Figure~\ref{fig:confusion}). The diagonal elements represent the recall for each game class. We observe strong diagonal performance for distinct representative games such as \textit{racebet} ($84\%$) and \textit{portals} ($66\%$).

However, consistent with the limitations in model understanding, confusions were largely contained within semantically similar clusters. For example, significant confusion exists between \textit{digdug} and \textit{boulderchase}. Both games involve grid-based digging and falling boulders, suggesting that while models can recognize general mechanical patterns from the ASCII grids, they struggle to distinguish the specific causal nuances and mechanics that define the exact game title. This may also be because the gameplay traces failed to capture the particular states and transitions that distinguish the different mechanisms.

\subsection{Task II: VGDL Synthesis Results}
\label{sec:results_task2}

We evaluated the fidelity of the generated Video Game Description Language (VGDL) mechanics using the dual-stream pipeline---Direct Synthesis vs. Structural Causal Model (SCM)-Guided---across 45 unique generation tasks. The evaluation compared \textbf{Qwen3-8B} against the quantized version of \textbf{QwQ-32B} model. Results were validated by a panel of four Large Language Model (LLM) judges (GPT-5.2, Gemini 2.5 Pro, Claude 4.5 Opus, and Qwen3-8B).

\subsubsection{Overall Model Performance and Judge Agreement}
As detailed in Table~\ref{tab:judge_wins}, the SCM-mediated approach (Stream B) was consistently preferred over Direct Synthesis (Stream A) across both models and all judges. 
For \textbf{Qwen3-8B}, the SCM stream achieved a dominant win rate of approximately \textbf{67\%--81\%}, with very few ties. 
In contrast, the larger \textbf{QwQ-32B} model showed a higher incidence of ties (up to 15.9\% with Opus), suggesting that its ``Direct'' synthesis is a little more often qualitatively closer to its ``SCM'' output than the smaller model's. However, the preference for SCM remains strong, peaking at \textbf{75\%} with the Qwen-based judge.

\vspace{1pt}
\begin{table}[!htbp]
\centering
\caption{Overall Judge Win Rates comparing Direct Synthesis (NS) vs. SCM-Guided (SCM). The abbreviation \textbf{NS} denotes ``Non-SCM'' (Direct Synthesis). The reasoning model (QwQ-32B) exhibits higher tie rates, but SCM remains the preferred method across all judges.}
\label{tab:judge_wins}
\resizebox{0.6\textwidth}{!}{
\begin{tabular}{|l|l|c|c|c|c|}
\hline
\textbf{Model} & \textbf{Judge} & \textbf{NS \%} & \textbf{SCM \%} & \textbf{Tie \%} \\
\hline
Qwen3-8B & GPT-5.2 & 31.0\% & \textbf{66.7\%} & 2.4\% \\
Qwen3-8B & Gemini 2.5 Pro & 33.3\% & \textbf{66.7\%} & 0.0\% \\
Qwen3-8B & Claude 4.5 Opus & 28.6\% & \textbf{66.7\%} & 4.8\% \\
Qwen3-8B & Qwen-3-8B & 19.0\% & \textbf{81.0\%} & 0.0\%  \\ \hline
QwQ-32B & GPT-5.2 & 22.2\% & \textbf{64.4\%} & 13.3\% \\
QwQ-32B & Gemini 2.5 Pro & 40.0\% & \textbf{60.0\%} & 0.0\% \\
QwQ-32B & Claude 4.5 Opus & 31.8\% & \textbf{52.3\%} & 15.9\% \\
QwQ-32B & Qwen-3-8B & 17.5\% & \textbf{75.0\%} & 7.5\% \\
\hline
\end{tabular}
}
\end{table}

\subsubsection{Context Level Sensitivity}
Table~\ref{tab:level_metrics} reveals a divergence in how the two models utilize context:

\begin{itemize}
    \item \textbf{Induction (Levels 0-3):} Both models benefit significantly from the SCM pipeline in low-information settings. At Level 0 (Observation Only), Qwen3-8B's SCM output was preferred \textbf{91.7\%} of the time. QwQ-32B also favored SCM (58.8\%), though with a high tie rate of 17.6\%, indicating that the larger model is better at implicit reasoning even without the explicit SCM step. 
    
    \item \textbf{The Completion Reversal (Level 4):} We observed a behavioral split at Level 4 (Partial VGDL). 
    \textbf{Qwen3-8B} reverted to Direct Synthesis, which won \textbf{62.5\%} of votes and achieved a higher Code Similarity (0.955 vs 0.939). This confirms that for smaller models, the overhead of generating a full SCM for simple completion tasks may introduce noise.
    Conversely, \textbf{QwQ-32B} maintained a preference for SCM (51.5\% wins) and achieved higher Code Similarity (0.909 vs 0.876) with the causal graph. This suggests that reasoning-optimized models can effectively leverage the SCM structure even when the task is primarily syntactic completion.
\end{itemize}
Also, the SCM-guided VGDL generation was consistently preferred across the benchmark games, except \textit{portals} in the case of QwQ-32B. A comprehensive breakdown of performance metrics across all games is provided in Table~\ref{tab:game_metrics} in Appendix~\ref{app:game_metrics}.

\begin{table}[!htbp]
\centering
\caption{Context Level Metrics. \textbf{NS} denotes Non-SCM (Direct Synthesis). Note the reversal at Level 4 for Qwen3-8B.}
\label{tab:level_metrics}
\resizebox{0.8\textwidth}{!}{%
\begin{tabular}{|l|l|c|c|c|c|c|c|c|}
\hline
\multirow{2}{*}{\textbf{Model}} & \multirow{2}{*}{\textbf{Target}} & \multicolumn{2}{c|}{\textbf{Non-SCM (NS)}} & \multicolumn{2}{c|}{\textbf{SCM-Guided}} & \multicolumn{3}{c|}{\textbf{Outcomes (Win \%)}} \\
\cline{3-9}
 & & \textbf{Code $\uparrow$} & \textbf{Sem $\uparrow$} & \textbf{Code $\uparrow$} & \textbf{Sem $\uparrow$} & \textbf{NS $\uparrow$} & \textbf{SCM $\uparrow$} & \textbf{Tie} \\
\hline
Qwen3-8B & L0: Obs Only & 0.563 & 0.736 & \textbf{0.653} & \textbf{0.740} & 8.3\% & \textbf{91.7\%} & 0.0\% \\
Qwen3-8B & L1: Grammar & 0.670 & 0.733 & \textbf{0.710} & \textbf{0.743} & 9.4\% & \textbf{84.4\%} & 6.2\% \\
Qwen3-8B & L2: Exact Game & 0.692 & 0.725 & \textbf{0.716} & \textbf{0.768} & 37.5\% & \textbf{59.4\%} & 3.1\% \\
Qwen3-8B & L3: Distractors & \textbf{0.694} & 0.714 & 0.693 & \textbf{0.755} & 25.0\% & \textbf{75.0\%} & 0.0\% \\
Qwen3-8B & L4: Partial & \textbf{0.955} & \textbf{0.895} & 0.939 & 0.871 & \textbf{62.5\%} & 37.5\% & 0.0\% \\
\hline
QwQ-32B & L0: Obs Only & 0.532 & 0.701 & \textbf{0.591} & \textbf{0.732} & 23.5\% & \textbf{58.8\%} & 17.6\% \\
QwQ-32B & L1: Grammar & \textbf{0.647} & 0.731 & 0.646 & \textbf{0.742} & 32.4\% & \textbf{52.9\%} & 14.7\% \\
QwQ-32B & L2: Exact Game & 0.660 & \textbf{0.790} & \textbf{0.661} & 0.760 & 11.8\% & \textbf{82.4\%} & 5.9\% \\
QwQ-32B & L3: Distractors & \textbf{0.659} & 0.752 & 0.656 & \textbf{0.764} & 26.5\% & \textbf{67.6\%} & 5.9\% \\
QwQ-32B & L4: Partial & 0.876 & 0.868 & \textbf{0.909} & \textbf{0.883} & 45.5\% & \textbf{51.5\%} & 3.0\% \\
\hline
% LEGEND ROW
\multicolumn{9}{|l|}{\footnotesize \textbf{Code}: VGDL Cosine Similarity; \textbf{Sem}: Semantic Text Similarity.} \\
\hline
\end{tabular}%
}
\end{table}

\section{Limitations, Future Work and Conclusion}
Our semantic clustering relied on text embeddings that may not have fully captured the mechanical nuances of the games, potentially resulting in overlapping clusters; future work should evaluate alternative embedding architectures and clustering algorithms. Computational constraints (maximum 48 GB VRAM) restricted our analysis to models with $\le$32B parameters and limited the synthesis task to a single observation set per game.

Current evaluation metrics for VGDL generation, such as simple parsability checks are often arbitrary and do not adequately measure causal induction. Developing robust metrics for causal fidelity is a critical next step. Finally, a ``belief SCM'' framework where an LLM iteratively updates an internal causal model as it processes sequential observations may be explored in the future. This approach would mitigate the high token overhead of batch processing while mimicking continuous learning.

These intermediate SCMs would also enable Causal Reinforcement Learning~\cite{deng2023causal}, facilitating interpretable AI and counterfactual reasoning. Moreover, perturbing these graphs advances Procedural Content Generation (PCG) by ensuring logically consistent mechanics, overcoming playability issues inherent to statistical methods~\cite{nielsen2015towards}.

In conclusion, we demonstrate that grounding Large Language Models in Structural Causal Models significantly improves the reverse-engineering of game mechanics from raw observations. Across diverse games, context regimes, and models, the SCM-guided pipeline consistently yields VGDL rules that are more accurate and logically consistent than direct synthesis alone, particularly in low-information and inductive settings. These results provide empirical evidence that explicitly structured causal reasoning offers a powerful inductive bias beyond purely statistical pattern matching, establishing a foundation for neurosymbolic agents that can more robustly infer, explain, and manipulate the rules governing their environments.

\subsubsection{Acknowledgments} 
The authors would like to thank Dr Diego Perez-Liebana for providing us with expert descriptions of the GVGAI games. The editing process was supported by Google's Gemini AI agent, which was used to suggest revisions for language, structure, and readability. ChatGPT was used to support the development of code for the experiments reported in this paper. 

\bibliographystyle{splncs04}
\bibliography{ICPR_2026_LaTeX_Templates/ref}

% -------------------------------------------------------------------
%                        Appendix
% -------------------------------------------------------------------

\newpage
\setcounter{page}{1}
\appendix
\section{Appendix}

\subsection{Example Observations with actions from the Game \textit{Realsokoban}}

For visual clarity, observations are illustrated using a compact game from the GVGAI environment, as the representative games' grids exceed the page layout constraints.

\begin{lstlisting}

wall   wall   wall   wall   wall   wall   wall   wall  
wall   wall                                      wall  
wall   wall          hole   box    hole          wall  
wall   wall          box    avatar        box    wall  
wall                 hole   box    hole          wall  
wall                                             wall  
wall   wall   wall   wall   wall   wall   wall   wall  

Action: Down

wall   wall   wall   wall   wall   wall   wall   wall  
wall   wall                                      wall  
wall   wall          hole   box    hole          wall  
wall   wall          box                  box    wall  
wall                 hole   avatar hole          wall  
wall                        box                  wall  
wall   wall   wall   wall   wall   wall   wall   wall  

Action: Up

wall   wall   wall   wall   wall   wall   wall   wall  
wall   wall                                      wall  
wall   wall          hole   box    hole          wall  
wall   wall          box    avatar        box    wall  
wall                 hole          hole          wall  
wall                        box                  wall  
wall   wall   wall   wall   wall   wall   wall   wall  

\end{lstlisting}
\newpage

\subsection{Preliminary Model Evaluation}

\begin{figure}[H]
    \centering
    \includegraphics[width=0.9\textwidth]{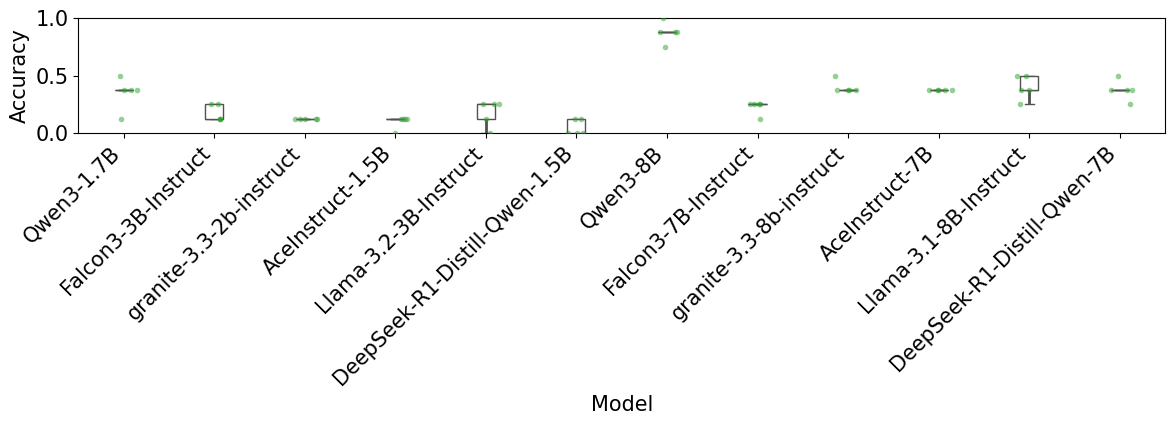}
    \caption{Preliminary accuracy assessment for a 10-way game classification task (using random games from~\cite{dockhorn2020prediction}). The Qwen3-8B model demonstrates significantly superior performance compared to other sub-4B and sub-8B models, justifying its selection for the VGDL translation task.}
    \label{fig:prelim_test}
\end{figure}

We evaluated varying models from cost-effective parameter categories (sub-4B and sub-8B) to ensure computational efficiency. As illustrated in Figure~\ref{fig:prelim_test}, the \textbf{Qwen3-8B} model emerged as a significant outlier compared to other tested architectures.

\subsection{Example natural language description generated by Qwen3-8B from VGDL of the game \textit{Brainman}}\label{app:vgdl_to_natural}

In \textit{Brainman}, the player controls an avatar navigating a maze. Collect keys (turning into missiles) to unlock doors and defeat enemies (gems) for points. Boulders bounce on contact, and walls/doors block movement. The avatar loses if killed by gems (1-5 points) or trapped. Winning requires reaching the exit (10 points). Key missiles destroy walls, while undo interactions reset objects. The game ends when the avatar dies or the exit is reached.

\subsection{Clustering Hyperparameters and Preprocessing}\label{sec:appendix_clustering}
\begin{figure}[H]
    \centering
    \includegraphics[width=\linewidth]{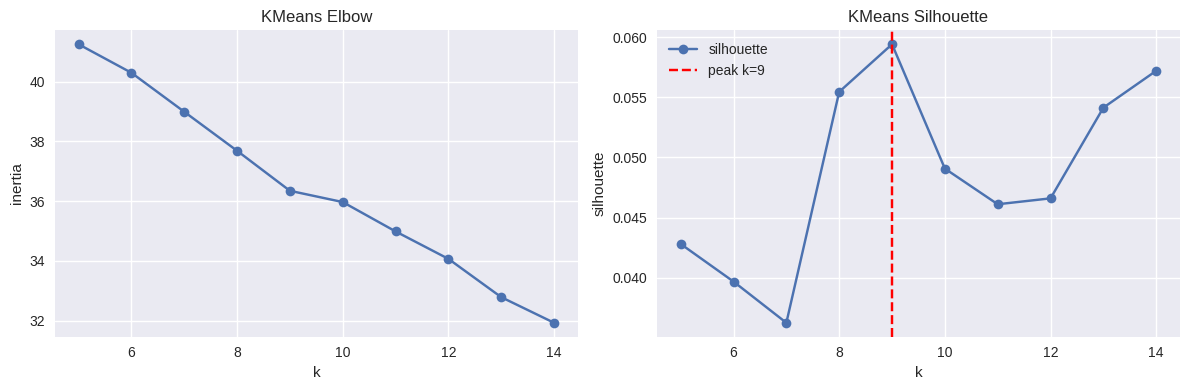}
    \caption{Evaluation of K-Means clustering metrics. The Elbow method (left) displays inertia reduction, while the Silhouette analysis (right) confirms $k=9$ as the optimal parameter.}
    \label{fig:kmeans_metrics}
\end{figure}

To determine the optimal number of semantic clusters, we performed a sweep over $k \in [5, 14]$, evaluating both \textit{Inertia} and \textit{Silhouette Score}. As shown in Figure~\ref{fig:kmeans_metrics}, the Silhouette analysis identifies a distinct global maximum at $k=9$.

Furthermore, we investigated the stability of these clusters under different preprocessing strategies. Although the official S-BERT example code implements K-Means directly on raw embeddings~\cite{sbert_kmeans_example}, we expanded our analysis to include Standard scaling~\cite{scikit-learn}, L2 normalization, and dimensionality reduction (PCA~\cite{PCA}, UMAP) for completeness.

\begin{figure}[H]
    \centering
    \includegraphics[width=\linewidth]{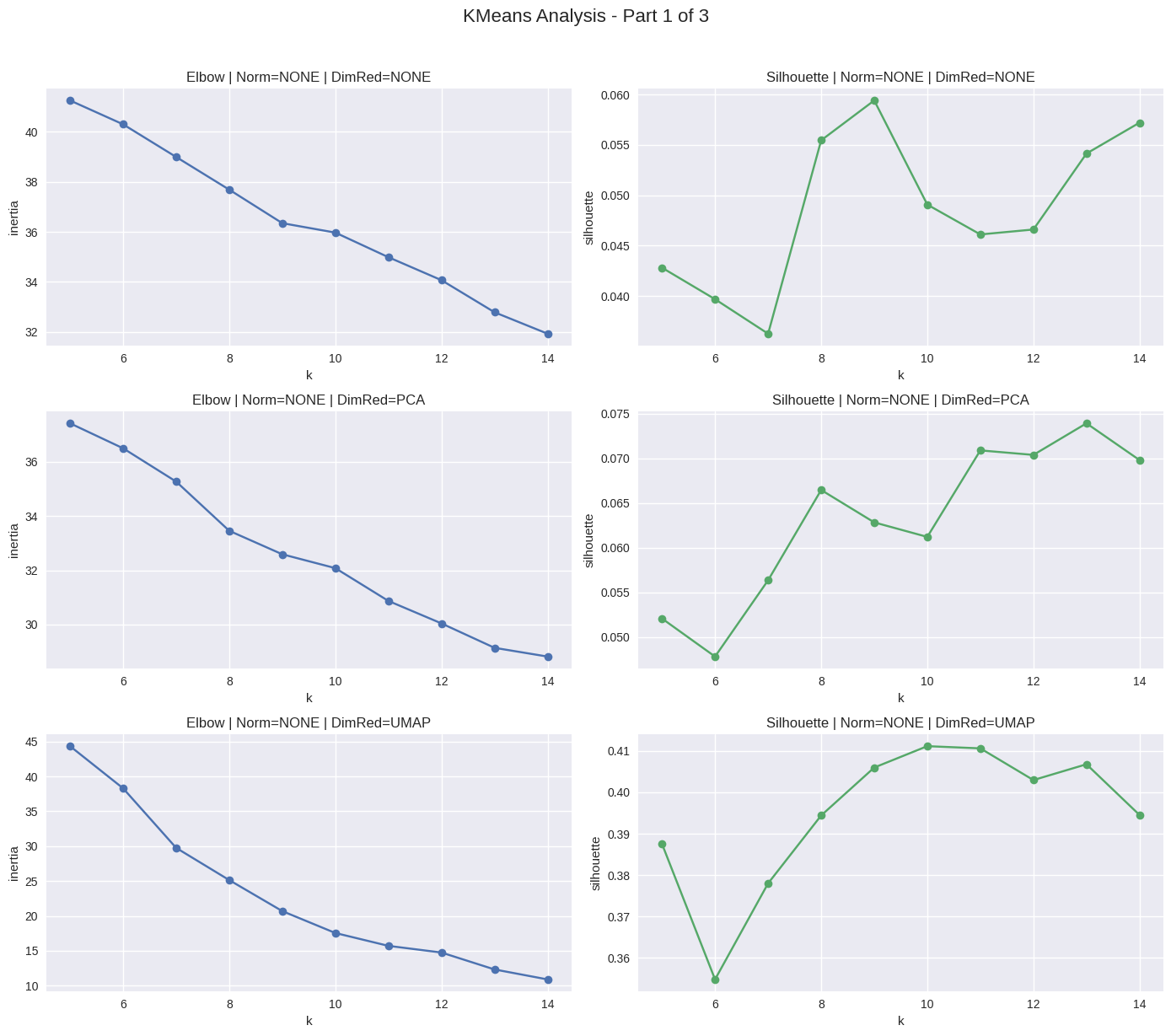}
    \caption{Clustering performance analysis (Part 1): Comparison of Inertia and Silhouette scores across different preprocessing combinations, including Standard scaling~\cite{scikit-learn} and L2 Normalization.}
    \label{fig:kmeans_analysis_1}
\end{figure}

\begin{figure}[H]
    \centering
    \includegraphics[width=\linewidth]{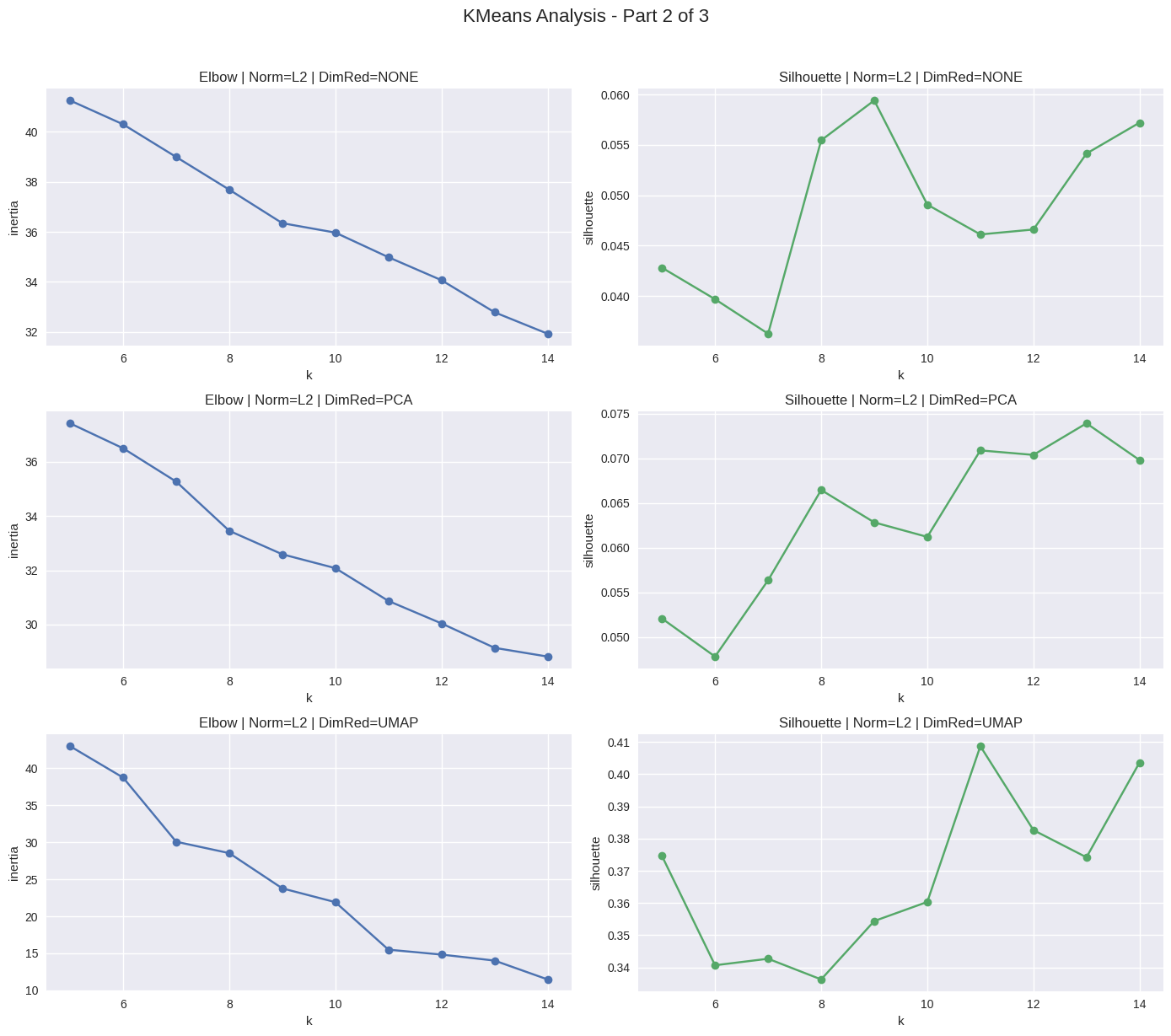}
    \caption{Clustering performance analysis (Part 2): Extended evaluation of dimensionality reduction techniques (PCA) combined with scaling strategies.}
    \label{fig:kmeans_analysis_2}
\end{figure}

\begin{figure}[H]
    \centering
    \includegraphics[width=\linewidth]{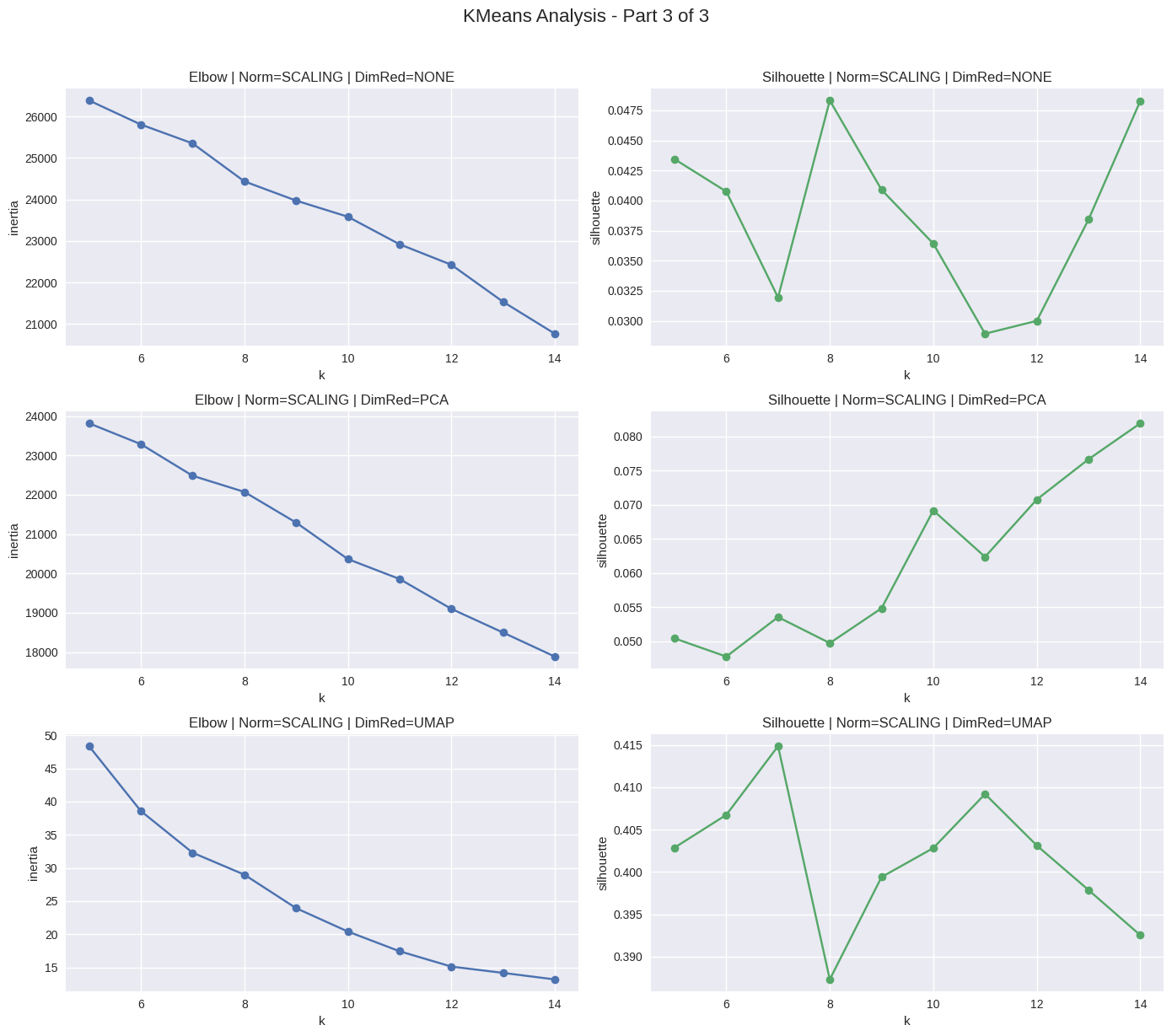}
    \caption{Clustering performance analysis (Part 3): Extended evaluation of manifold learning techniques (UMAP) combined with scaling strategies.}
    \label{fig:kmeans_analysis_3}
\end{figure}

Figures~\ref{fig:kmeans_analysis_1}--\ref{fig:kmeans_analysis_3} illustrate the Elbow and Silhouette metrics across these exhaustive configurations. We observed that while dimensionality reduction and scaling can aid in visualization, the raw, 384-dimensional embeddings provided robust cluster separation without the information loss associated with projection, validating our decision to follow the official documentation's approach for the main analysis.

\subsection{Clusters made using K-means clustering}\label{sec:clusters_extended}

\begin{longtable}{|l|l|p{0.5\textwidth}|}
% Caption and label for the table
\caption{Extended KMeans clustering results with all cluster members.}
\label{tab:kmeans_clusters_extended}
\\

% --- Table Header for the first page ---
\hline
\textbf{Cluster ID} & \textbf{Representative game} & \textbf{Members} \\
\hline
\endfirsthead

% --- Table Header for all subsequent pages ---
\multicolumn{3}{c}%
{{\tablename\ \thetable\ -- \textit{Continued from previous page}}} \\
\hline
\textbf{Cluster ID} & \textbf{Representative game} & \textbf{Members} \\
\hline
\endhead

% --- Table Footer for all pages except the last ---
\hline \multicolumn{3}{r}{{\textit{Continued on next page}}} \\
\endfoot

% --- Table Footer for the last page ---
\hline
\endlastfoot

% --- Table Data ---
00 & racebet     & racebet, camelRace, racebet2 \\ \hline
01 & digdug      & superman, islands, rivers, chase, plants, frogs, eggomania, digdug, plaqueattack, painter, sheriff, pacman, gymkhana, surround, bait, overload, whackamole, seaquest \\ \hline
02 & brainman    & brainman, roguelike, escape, labyrinthdual, intersection, realsokoban, avoidgeorge, lemmings, run, freeway, enemycitadel, blacksmoke, labyrinth, sokoban, zelda, thecitadel, clusters, waitforbreakfast, cops, dungeon \\ \hline
03 & defender    & defender, waves, aliens, missilecommand, solarfox \\ \hline
04 & lasers2     & boloadventures, assemblyline, lasers2, factorymanager, lasers \\ \hline
05 & boulderchase & shipwreck, boulderdash, jaws, boulderchase, chainreaction, butterflies \\ \hline
06 & portals     & portals, colourescape, crossfire, witnessprotection, zenpuzzle, realportals, chopper, catapults, tercio, bomber, infection, modality, angelsdemons \\ \hline
07 & iceandfire  & chipschallenge, survivezombies, firestorms, thesnowman, iceandfire, fireman, firecaster \\ \hline
08 & cakybaky    & cookmepasta, hungrybirds, cakybaky \\ \hline

\end{longtable}

\subsection{Game descriptions provided by expert}

\textbf{Racebet:} The objective of this game is guessing which camel is going to win the race. Betting is performed by placing the player in one of the corresponding betting spots. No points are given in this game. You either bet for the one that wins (and the player wins) or you don't (and the player loses).

\textbf{Digdug:} The player must collect all gems and gold coins in the cave, digging its way through it, to win the game. There are also enemies in the level that kill the player on collision with them. Also, the player can shoot boulders by pressing USE two consecutive time steps, which kill enemies.

\textbf{Brainman:} The player must push keys into doors to open them, to reach the exit to win the game. Many gems around are available for collection, which give score to the player. Keys are pushed by colliding with them, and they are propelled in a straight line until finding a sprite to collide against, when they stop. If they hit a door, both the door and the key are destroyed.

\textbf{Defender:} The objective of this game is to avoid that the enemies destroy your city. The enemies are spawned at the right of the screen and cross it while bombing the city. The player must eliminate these enemies as soon as possible so the city is not destroyed, otherwise the player loses. The player’s ammunition is supplied from the top of the screen and must be collected periodically to avoid running out of ammo.

\textbf{lasers2:} The objective of this game is to reach the exit, which is typically blocked by a mud sprite. The mud can be destroyed by directing a laser to it. There are several laser cannons whose orientation can be changed by the player shooting at them. Apart from this, there are several crystal blocks, which can also be rotated, that deflect the lasers in a certain direction. Lasers kill the player when in contact.

\textbf{Boulderchase:} In this game, the player must dig inside a cave to collect at least 9 gems and reach the exit. As the player digs, boulders that are left without a ground fall and can kill the player upon collision. Enemies also move around the cave, being able to dig as well. When digging, they drop gems that can also be collected by the player. Enemies are also killed by boulders and kill the player upon contact.

\textbf{Portals:} The player controls an avatar that needs to find the exit of a maze. The maze is full of entry and exit portals, and every time the player goes through an entry portal they will be teleported to an exit portal at random. The maze is also full of enemies and missiles that kill the player on contact.

\textbf{Iceandfire:} The objective of this game is to find the exit of the maze. The maze has multiple traps, two different types of surfaces (ice and fire), and two types of boots (for ice and fire) that can be collected. The special surfaces kill the player on contact, unless the appropriate boots are picked up beforehand. There are several coins scattered around the level that give the player some score.
\newpage

\subsection{Example of a ground-truth VGDL for the game \textit{Brainman}}
\label{app:vgdl}

\begin{lstlisting}
BasicGame
    SpriteSet
        floor > Immovable hidden=True img=newset/floor2
        avatar > OrientedAvatar img=oryx/prince1 rotateInPlace=false
        gem > Immovable shrinkfactor=0.7
            green >  img=oryx/diamond3
            red >  img=oryx/diamond2
            blue >  img=oryx/diamond1
        key > Passive img=oryx/key3 shrinkfactor=0.7
        keym > Missile img=oryx/key3 shrinkfactor=0.6
        exit > Immovable img=newset/exit2
        door > Immovable img=oryx/doorclosed1
        boulder > Passive img=newset/block3
        wall > Immovable img=oryx/wall3 autotiling=True

    LevelMapping
        A > avatar floor
        k > key floor
        d > door floor
        e > exit floor
        g > green floor
        r > red floor
        b > blue  floor
        O > boulder floor
        . > floor

    InteractionSet
        keym key wall gem boulder > transformTo stype=key
        avatar wall door > stepBack        
        boulder avatar > bounceForward        
        key avatar > transformTo stype=keym
        keym avatar > attractGaze
        avatar key keym > stepBack        
        door keym > killBoth scoreChange=4
        green avatar > killSprite scoreChange=1
        blue avatar > killSprite scoreChange=2
        red avatar > killSprite scoreChange=5        
        boulder wall key gem boulder > undoAll        
        key wall gem key > undoAll
        exit avatar > killSprite scoreChange=10
        
    TerminationSet
        SpriteCounter stype=avatar limit=0 win=False
        SpriteCounter stype=exit limit=0 win=True
\end{lstlisting}

\subsection{SCM blueprint}

\begin{figure}[!htbp]
\begin{lstlisting}[language=Python, caption={SCM Blueprint Definition}, label={lst:scm_blueprint}]
nodes={
    # Design Layer
    "EntityTypes": "Static sprite types, roles (avatar, wall), & attrs (speed, hp). Not time-varying.",
    "ActionSpace": "Agent interventions/actions available at each step (move, shoot).",
    "GlobalMechanics": "Global state transitions independent of collisions (gravity, timers).",
    "InteractionMechanics": "Collision rules mapping (Entities, State_t, Action_t) -> State_{t+1} events (spawn, destroy).",
    "RewardMechanics": "Maps State_t/Interactions to Reward_t or Score_{t+1}.",
    "TerminationMechanics": "Win/loss conditions based on StateVariables (e.g., timeout, goals_met).",
    # Dynamics Layer
    "StateVariables": "Dynamic vars evolving over t (pos, hp). Eq: X_{t+1}=f(PA_X_t, Action_t, U_X).",
    "InitialState": "StateVariables at t=0 derived from EntityTypes + LevelEncoding.",
    # Observation Layer
    "LevelEncoding": "ASCII grid mapping chars to entity instances/positions; induces InitialState.",
},
edges=[
    ("EntityTypes", "InitialState"), ("LevelEncoding", "InitialState"), ("EntityTypes", "StateVariables"),
    ("ActionSpace", "InteractionMechanics"), ("ActionSpace", "StateVariables"),
    ("GlobalMechanics", "StateVariables"), ("InteractionMechanics", "StateVariables"),
    ("StateVariables", "InteractionMechanics"), ("StateVariables", "RewardMechanics"),
    ("InteractionMechanics", "RewardMechanics"), ("RewardMechanics", "StateVariables"),
    ("StateVariables", "TerminationMechanics"), ("InteractionMechanics", "TerminationMechanics"),
],
scm_notes=[
    "Define DYNAMIC SCM with discrete time steps t=0,1...",
    "Separate STATIC design vars from DYNAMIC state vars.",
    "For dynamic X, define X_{t+1} = f(Parents_t, Action_t, Noise).",
    "Define Reward_t & Termination explicitly as equations.",
    "Output machine-readable format (JSON) with: static_nodes, dynamic_variables, edges, equations.",
    "Link LevelEncoding chars to InitialState.",
],
vgdl_notes=[
    "EntityTypes+StateVariables -> VGDL SpriteSet (roles, attrs).",
    "LevelEncoding+InitialState -> VGDL LevelMapping.",
    "InteractionMechanics+GlobalMechanics -> VGDL InteractionSet.",
    "RewardMechanics+TerminationMechanics -> VGDL TerminationSet.",
    "ActionSpace -> VGDL Avatar controls.",
],
\end{lstlisting}
\end{figure}

\subsection{Game-Specific Analysis for Task II}
\label{app:game_metrics}

Table~\ref{tab:game_metrics} details the performance breakdown by game, highlighting the specific code and semantic similarity scores for both generation streams.

Notably, the \textit{portals} environment presented a distinct challenge, resulting in a \textbf{90\% tie rate} for QwQ-32B. This was largely due to the extensive length of the game's observations, which strained the model's generation capabilities. As a result, the model frequently struggled to produce executable VGDL in either condition. The high tie rate, therefore, reflects a shared difficulty in handling this heavy token load, rather than an equivalence in successful reasoning.

\begin{table}[H]
\centering
\caption{Game-Specific Metrics. Comparison of Code and Semantic similarities across key games for Non-SCM (NS) and SCM-Guided (SCM) streams.}
\label{tab:game_metrics}
\resizebox{\textwidth}{!}{
\begin{tabular}{|l|l|c|c|c|c|c|c|c|}
\hline
\textbf{Model} & \textbf{Game} & \textbf{NS-Code $\uparrow$} & \textbf{NS-Sem $\uparrow$} & \textbf{SCM-Code $\uparrow$} & \textbf{SCM-Sem $\uparrow$} & \textbf{NS Win\% $\uparrow$} & \textbf{SCM Win\% $\uparrow$} & \textbf{Tie\%} \\
\hline
Qwen3-8B & boulderchase & 0.788 & 0.777 & \textbf{0.816} & \textbf{0.783} & 31.2\% & \textbf{68.8\%} & 0.0\% \\
Qwen3-8B & brainman & 0.664 & 0.775 & \textbf{0.668} & \textbf{0.788} & 15.0\% & \textbf{85.0\%} & 0.0\% \\
Qwen3-8B & cakybaky & \textbf{0.747} & \textbf{0.801} & 0.739 & 0.791 & 25.0\% & \textbf{75.0\%} & 0.0\% \\
Qwen3-8B & defender & 0.631 & 0.656 & \textbf{0.646} & \textbf{0.679} & 30.0\% & \textbf{70.0\%} & 0.0\% \\
Qwen3-8B & digdug & 0.720 & \textbf{0.806} & \textbf{0.764} & \textbf{0.806} & 35.0\% & \textbf{60.0\%} & 5.0\% \\
Qwen3-8B & iceandfire & 0.766 & 0.743 & \textbf{0.797} & \textbf{0.776} & 37.5\% & \textbf{62.5\%} & 0.0\% \\
Qwen3-8B & lasers2 & 0.647 & 0.757 & \textbf{0.741} & \textbf{0.809} & 20.0\% & \textbf{75.0\%} & 5.0\% \\
Qwen3-8B & portals & 0.767 & \textbf{0.725} & \textbf{0.786} & 0.721 & 25.0\% & \textbf{75.0\%} & 0.0\% \\
Qwen3-8B & racebet & 0.703 & 0.807 & \textbf{0.722} & \textbf{0.820} & 35.0\% & \textbf{60.0\%} & 5.0\% \\ \hline
QwQ-32B & boulderchase &\textbf{ 0.815} & 0.795 & 0.802 & \textbf{0.826} & 40.0\% & \textbf{55.0\%} & 5.0\% \\
QwQ-32B & brainman & 0.675 & \textbf{0.796} & \textbf{0.727} & 0.750 & 35.0\% & \textbf{65.0\%} & 0.0\% \\
QwQ-32B & cakybaky & 0.720 & 0.775 & \textbf{0.756} & \textbf{0.778} & 10.0\% & \textbf{90.0\%} & 0.0\% \\
QwQ-32B & defender & 0.609 & 0.705 & \textbf{0.673} & \textbf{0.739} & 40.0\% & \textbf{60.0\%} & 0.0\% \\
QwQ-32B & digdug & 0.740 & \textbf{0.789} & \textbf{0.755} & \textbf{0.789} & 20.0\% & \textbf{75.0\%} & 5.0\% \\
QwQ-32B & iceandfire & 0.757 & 0.797 & \textbf{0.761} & \textbf{0.843} & 45.0\% & \textbf{50.0\%} & 5.0\% \\
QwQ-32B & lasers2 & 0.684 & \textbf{0.818} & \textbf{0.775} & 0.752 & 15.0\% & \textbf{75.0\%} & 10.0\% \\
QwQ-32B & portals & \textbf{0.367} & 0.591 & 0.245 & \textbf{0.649} & 0.0\% & 10.0\% & \ \textbf{90.0\%} \\
QwQ-32B & racebet & 0.706 & 0.850 & \textbf{0.738} & \textbf{0.858} & 31.6\% & \textbf{57.9\%} & 10.5\% \\
\hline
\end{tabular}
}
\end{table}

\end{document}